\documentclass[letterpaper, 10 pt, conference]{ieeeconf}  % Comment this line out if you need a4paper
\IEEEoverridecommandlockouts                              
\usepackage{amsmath,graphicx,hyperref,subcaption} % assumes amsmath package installed
\usepackage[noadjust]{cite}
\usepackage{graphicx}
\usepackage{subcaption,xcolor}
\renewcommand{\theenumi}{\roman{enumi}}%

\title{\LARGE \bf
Design Iterations for Passive Aerial Manipulator
}
\author{Vidyadhara B. V.$^{1}$, Lima Agnel Tony$^{1}$, Mohitvishnu S. Gadde$^{1}$, Shuvrangshu Jana$^{1}$,\\
 Varun V. P.$^{2}$, Aashay Anil Bhise$^{1}$, Suresh Sundaram$^{3}$, Debasish Ghose$^{1}$% <-this % stops a space

\thanks{*This work is partially supported by IISc and Robert Bosch Center for Cyber Physical Systems, (IISc) and Khalifa University}% <-this % stops a space
\thanks{$^{1}$Guidance, Control, and Decision Systems Laboratory (GCDSL), Department of Aerospace Engineering,
        Indian Institute of Science, Bangalore-12, India.
        {\tt\small vidyadhara.vk@gmail.com, limatony@iisc.ac.in, mohitvishnug@iisc.ac.in,  shuvrangshuj@iisc.ac.in, meetaashay3@gmail.com, dghose@iisc.ac.in}}%
\thanks{$^{2}$
        Robert Bosch Center for Cyber Physical Systems, Bangalore-12, India.
        {\tt\small varunvp@iisc.ac.in}}%
\thanks{$^{3}$ Artificial Intelligence and Robotics Laboratory (ARL), Department of Aerospace Engineering,
        Indian Institute of Science, Bangalore-12, India.
        {\tt\small vssuresh@iisc.ac.in}}%
}

\begin{document}
\maketitle
\thispagestyle{empty}
\pagestyle{empty}

%%%%%%%%%%%%%%%%%%%%%%%%%%%%%%%%%%%%%%%%%%%%%%%%%%%%%%%%%%%%%%%%%%%%%%%%%%%%%%%%
\begin{abstract}
Grabbing a maneuvering target using drones is a challenging problem. This paper presents the design, development, and prototyping of a novel aerial manipulator for target interception. It is a single Degree of Freedom (DoF) manipulator with passive basket-type end-effector. The proposed design is energy efficient, light weight and suitable for aerial grabbing applications. The detailed design of the proposed manipulation mechanism and a novel in-flight extending propeller guard, is reported in this paper. 
\end{abstract}

\section{Introduction}\label{s0}
%%%% To be edited
Robotic arms on drones could automate a variety of mundane and tedious tasks performed in less reachable and dangerous settings. Few of these include facade maintenance of buildings,  communication tower inspection and repair, fire fighting, load transport, etc. In each of these applications, the design of the robotic arm or the manipulator is specific to the problem being addressed. The design of manipulator end-effector is based on the tools that it needs to work with to execute the desired task. Irrespective of the requirement, the limitations in the existing technology imposes many constraints on the manipulators like being energy optimal, low weight, etc., to perform the mission properly. Thus, design and development of aerial manipulators require a through consideration in various directions for successful execution of the task.

Several works on design of aerial manipulators for varied applications are found in literature. Design of mechanism and electronics involved of a contact inspection manipulator is given in \cite{fumagalli2016mechatronic}. Dexterous manipulator design for operations like wrenching is given in \cite{6523891}. Design and analysis of a parallel manipulators are presented in detail, in \cite{7219682,7759715}. Design of a six DoF manipulator, for bar structure assembly is given in \cite{cano2013mechanical}. Design and prototyping of a serial multi-link and compact manipulator is presented in \cite{7536029}. Design of a five bar mechanism with omni-directional workspace is given in \cite{9144837}. A survey of aerial manipulator modelling and control is given in \cite{meng2020survey}. \cite{nedungadi2019design} presents an active manipulator design for pick and place operations. Design of manipulation mechanisms for aerial grasping are presented in \cite{8278328,yu2019exploring,paul2020tams}.  Most of these  focuses on the modelling of the final design, and the control of the integrated system. It would be interesting to look into the manipulator design from its idea form to the final prototype, which would bring out the intricate details on its developmental aspects.

In this work, we present the design and modelling, from scratch, of an aerial manipulation system for moving target grabbing. The proposed end-effector design is novel and simple passive basket-type manipulator. The final design can be used to detach any moving target, up to 150 g weight and 20 cm in size (diameter), for the dimensions and choice of material of the end-effector prototype. The design and prototyping of a self extending propeller guard is also presented, which ensures drone safety during grabbing. The proposed manipulation mechanism can be employed for many applications from automated fruit picking in large orchards to counter-drone missions in defense applications. The design is energy optimal and could be manufactured easily and cost effectively. Safety is also a major aspect when it comes to the aerial grabbing of moving target. This paper also presents in detail the design and prototyping of an in-flight extending drone guard, which protects the propeller during the grabbing process.

The paper is organised as follows: Section \ref{sec:2} describes the problem statement along with the challenges present in achieving the task. Section \ref{sec:3} gives the detailed design of the manipulator end-effector, manipulator arm, and safety accessories.  Section \ref{sec:7} discusses the possible improvements on the final design. Section \ref{sec:8} concludes the paper.

\section{Motivation and Challenges Involved}\label{sec:2}
The problem of moving target grabbing is inspired from challenge 1 of MBZIRC 2020 \cite{MBZ}. The target to be grabbed is a ball which is suspended below another drone moving in some fashion. The ball is attached to the other drone via a flexible rod and the attachment between the ball and the rod is magnetic. The success of mission would involve detachment of the ball from the other drone and depositing it in a box. As mentioned in previous section, this problem in general could cater a wide range of applications. Capturing a maneuvering target requires solution to many interesting challenges.

The main challenges involved in the design and development of the manipulator for this problem, are listed below.
\begin{enumerate}
    \item Proper choice of manipulator location with respect to drone body, considering stability during engagement.
    \item Optimal extension of end-effector from drone to reach the ball, without compromising safety and stability.
    \item Suppress uncontrolled vibrations.
    \item Structural strength, to apply desired detachment force to remove ball.
    \item Remain within a net volume of 120 cm $\times$ 120 cm $\times$ 50 cm during take-off and landing phase, keeping the size of integrated system bounded.
\end{enumerate}

\section{Manipulator Design}\label{sec:3}
The following design iterations are followed to achieve a robust design. 
\subsection{End effector}
The end effector design requires certain essential capabilities like quick response, low weight, optimal power consumption, etc. The following designs were tested to develop a robust end effector for the considered ball grabbing problem.
\subsubsection{Interlocking claws}
\begin{figure}[htb!]
	\centering
	\includegraphics[width=1\linewidth]{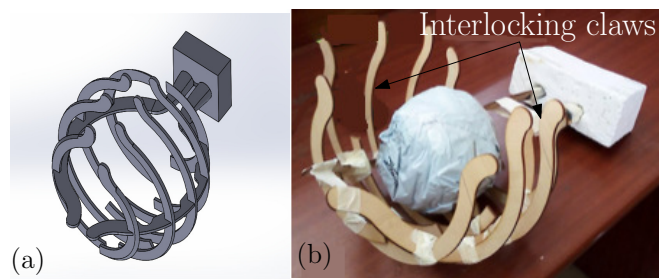}  
    \caption{Active interlocking claw (a) CAD design (b) Prototype}
    \label{fig:c1_EF1}
\end{figure}

With careful considerations of the challenges mentioned in Section \ref{sec:2}, a suitable manipulator end-effector was conceived. The primary design was a bio-inspired one, similar to that of how humans would catch a ball. The CAD design and its prototype, are shown in Fig. \ref{fig:c1_EF1}(a),(b), respectively. 

This design was primarily focused on the requirement of the end-effector to be within the size constraints and to be able drop the ball. This proof of concept was to be tested on a smaller drone and hence did not need any actuation. The prototype was fabricated using birch wood. The gripper was designed with two interlocking claws, which would snap into a locked position to grasp the ball. The snapping was actuated by a servo at the joining wrist. To drop the ball, the claws opened in the downward direction. Initial prototyping and testing showed several deficiencies in conceptualization: the success of grabbing was  dependant on the actuation timing of the servo. The delay posed a problem. Also, the energy budget had to be analysed thoroughly. Though the energy consumption for a servo is low, the servo needed to be engaged and disengaged till the ball was grabbed.  From the design perspective, the end-effector weight had detrimental effects on the servo's functionality: a large sized manipulator was not feasible due to its increased weight, which needed a larger servo for higher nominal torque, increasing the energy requirement. The observations led to the conclusion that an end-effector of passive nature would be better suited for the job and multiple DoF are more difficult to handle due to the complexity of control and associated delays.

\subsubsection{Passive basket}
The importance of endurance was understood from the initial tests with the active manipulator discussed above. Hence, the requirement of using a passive end-effector with minimal energy usage was incorporated into the next iterations.

The redesign of the manipulator was inspired by the fruit picking mechanism used in orchards, where a metallic wire is used to pluck the fruit which is collected in a net under the wire. The grasping problem could be visualised in a similar way with many added benefits. The mechanism was passive and thus energy needed was low, provided a larger weight margin due to absence of servos and additional wiring. This gave room for increasing the size of the end-effector. The drawback was that, a way to detach the ball was absent, since no interlocking claws were present. 

\begin{figure}[htb!]
	\centering
	\includegraphics[width=1\linewidth]{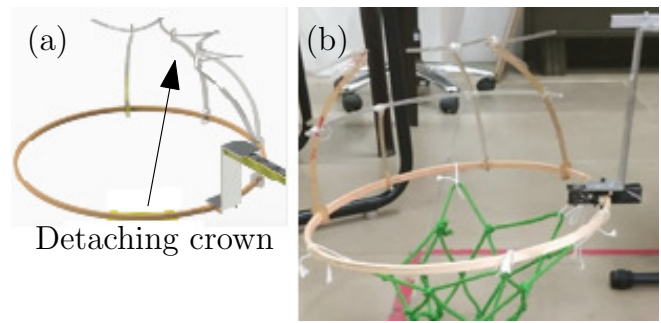}
    \caption{Passive basket with crown (a) CAD design (b) Prototype}
    \label{fig:c1_EF2}
\end{figure}

\begin{figure}[htb!]
    \centering
    \includegraphics[scale=0.75]{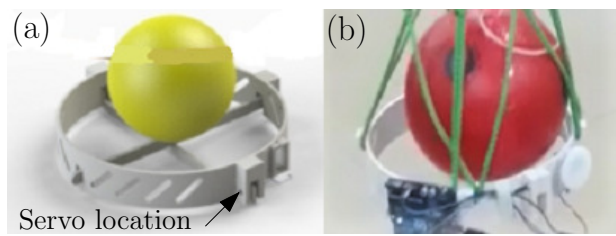}
    \caption{Dropping mechanism (a) CAD model (b) Prototype }
    \label{c1-drop}
\end{figure}

The first design among the passive manipulator is shown in Fig. \ref{fig:c1_EF2}(a). It fulfilled the energy requirement and the size constraints along with optimal weight and ball dropping capabilities. It had a crown for the ball detachment. The prototype was realised using wooden ring and acrylic strips (Fig. \ref{fig:c1_EF2}(b)). A net was fixed below for collecting the detached ball. A ring was fixed to the lower end of the net, which could be actuated to drop the ball, as shown in Fig. \ref{c1-drop}. 

\begin{figure}[htb!]
	\centering
	\includegraphics[width=1\linewidth]{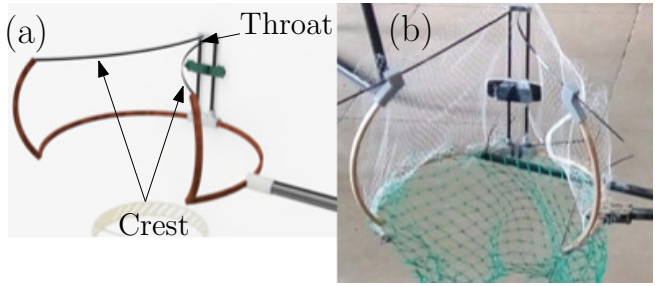}
    \caption{Modified passive basket with crest (a) CAD model (b) Prototype}
    \label{fig:c1_EF3}
\end{figure}

\begin{figure}[htb!]
	\centering
	\includegraphics[width=1\linewidth]{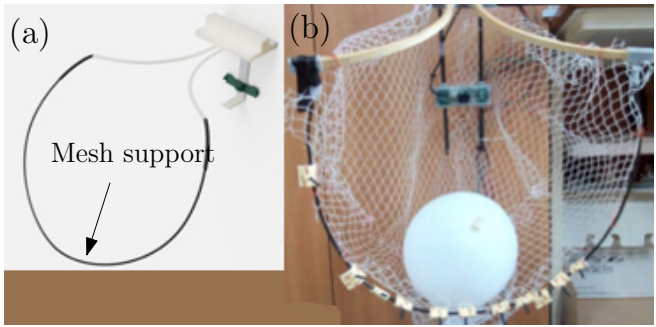}
    \caption{Modified basket with mesh support (a) CAD design (b) Prototype}
    \label{fig:c1_EF4}
\end{figure}

Testing of the new design brought in new issues. The wooden ring of the basket was hitting the ball to be grabbed and thus leading to failed attempts. From the structural point of view, the acrylic crown was prone to breaking after few attempts of grabbing. The design required an 'eye-in-hand' configuration for visual feedback. The positioning of camera in the crown was also found to be a difficult task without repeated calibrations. All these helped to arrive at a better design, shown in Fig. \ref{fig:c1_EF3}(a). 
A dedicated mount was erected for the camera. Instead of multiple detachment points, the crown was redesigned to a crest with a throat which acted as the single detachment point. Anything caught in the grabbing volume was naturally led to this point due to the relative motion of the gripper and ball.  The set-up was mounted at an elevation from the camera, aiding unobstructed feedback. The prototype is shown in Fig. \ref{fig:c1_EF3}(b). A slight modification led to the next design where the sagging net in the previous design was well supported and kept out of field of view of the camera (Fig. \ref{fig:c1_EF4}(a),(b)).
The experiments showed the need for larger grab volumes to accommodate errors in detection. The final prototype was designed to fulfil all the requirements of the end-effector.

\begin{figure}[htb!]
	\centering
    \includegraphics[width=1\linewidth]{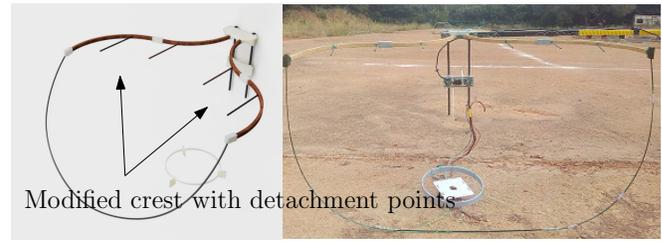}
    \caption{(a) Model of the passive basket gripper (b) Prototype }
    \label{fig:c1_EF5}
\end{figure}

The final prototype incorporated a convex nozzle design using two prongs made from birch wood reinforced with resin. The simplistic design was effective as it directed the ball from any point towards the throat, as shown in Fig. \ref{fig:c1_EF5}(a). In this design, to increase redundancy at the extreme ends of the prongs, additional disengaging probes were included. This was based on the experimental observations, where it was noticed that the end-effector could not direct the ball when contacted at these ends, thus requiring these inclusions to ensure success.  The net was hand woven from nylon thread, to minimize weight and also to reduce air drag (Fig. \ref{fig:c1_EF5}(b)).

\begin{figure}[htb!]
	\centering
	\includegraphics[scale=0.75]{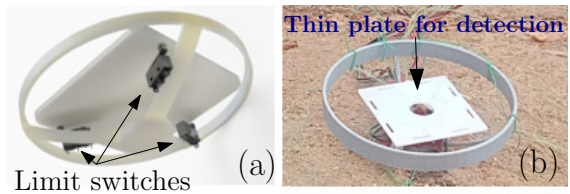}
    \caption{Grab detector (a) Model (b) Prototype}
    \label{c1-grab-det}
\end{figure}

A ring was fixed to the lower end of the net, which could be actuated to drop the ball, as shown in Fig. \ref{c1-grab-det}(a). An important addition to the design was that of a grab detector. Adding a vision feedback was an obvious choice but at the cost of increased energy and computational load. A thin plate detector was designed with three switches placed circularly with a plate to improve response sensitivity. The set-up was calibrated to actuate upon the ball falling on it (Fig. \ref{c1-grab-det}(b)). The dropping capability was ensured to be functional despite this addition.

\subsection{Manipulator arm}
\begin{figure}[htb!]
	\centering
	\begin{subfigure}{0.45\columnwidth}
	    \centering
        \includegraphics[scale=0.35]{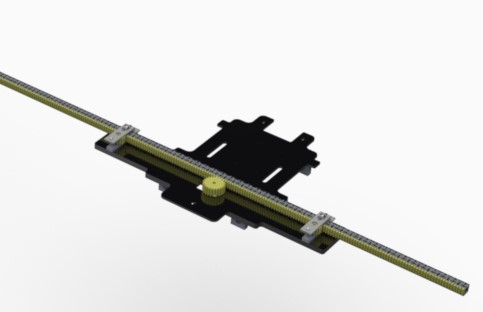}
        \vspace{1.25cm}
	    \subcaption{}
	\end{subfigure}
	\begin{subfigure}{0.5\columnwidth}
	    \centering
	    \includegraphics[scale=0.55]{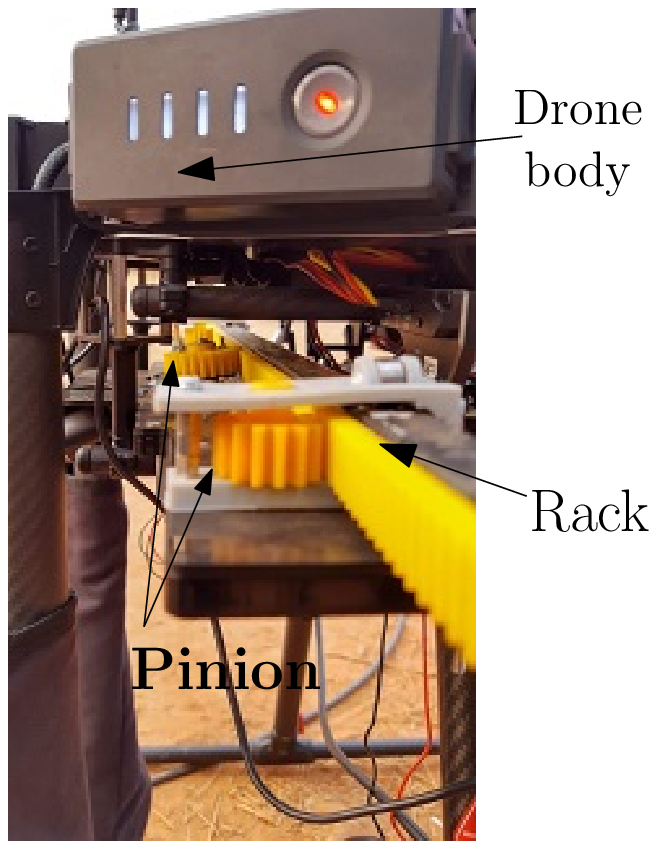}
	    \subcaption{}
	\end{subfigure}
    \caption{(a) CAD model of the rack and pinion arrangement with its mount (b) Side view of the on-drone set up}
    \label{rck}
\end{figure}

The manipulator arm for the end-effector was designed to have a single degree of freedom. The orientation is kept sideways, the intuitive reason being safety and effective grabbing using relative motion. An extension-retraction mechanism was implemented with a simple and effective rack and pinion set-up actuated by a continuous rotating servos, as shown in Fig. \ref{rck}(a). To tighten the grip and hence reduce sag, idler pinions were added on either sides of the actuating pinion (Fig. \ref{rck}(b)). Feedback was necessary at different stages of the mission: to know if the manipulator has extended to the complete extent to start the mission and to know if the manipulator is completely retracted at the end of the mission. For all such feedback requirements, limit switches were used. The choice of material for the arm is made in such a way that structural strength and minimal vibrations are ensured.
\subsection{Drone}
\begin{figure}[htb!]
	\centering
	\includegraphics[width=1\linewidth]{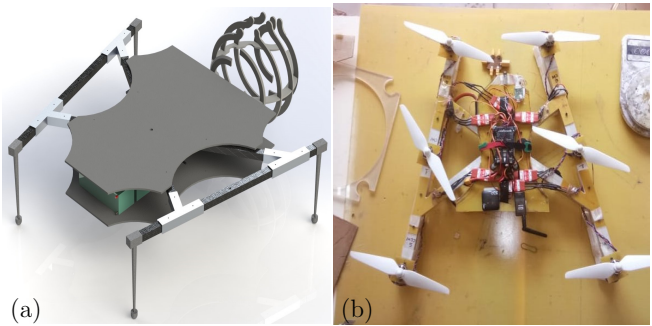}
    \caption{Hexa copter design for the interlocking claw manipulator (a) CAD design (b) Prototype}
    \label{drone_design}
\end{figure}

At the initial stage, a hexacopter was designed and prototyped, to incorporate the end-effector design within the specified volume constraint, as shown in Fig. \ref{drone_design}(a).  
%%%Added
The hexacopter was designed with a V configuration with an intention to mount the manipulator in the wide end as shown in the CAD model. This arrangement was chosen as it would ensure that the manipulator was not affected by the downwash, and would contain the claw design in its body front. Hence, the mechanism did not need an extension away from the drone frame. The prototype built (Fig. \ref{drone_design}(b)) was test flown to confirm stable flight.
The V configuration posed issue with maximum propeller size that can be used, along with the difficulty in calibration due to its unconventional configuration. The limit on propeller size limited the maximum thrust. Hence the prototype had low payload capacity making integration of manipulator difficult.  
The material used for fabricating the drone frame is Aluminium. For the size used, the strength to weight ratio is lower than a composite construction. By using custom made carbon fiber tubes, lighter and sturdier frame could be built. Using better motors could improve the thrust for the same propeller size though this may be a costlier investment. Modifying the end-effector to passive basket, the requirement of a manipulator arm made the current design infeasible due to stability issues and larger volume requirements. So, further tests were carried out using a different platform.

\begin{figure}[htb!]
	\centering
	\includegraphics[width=1\linewidth]{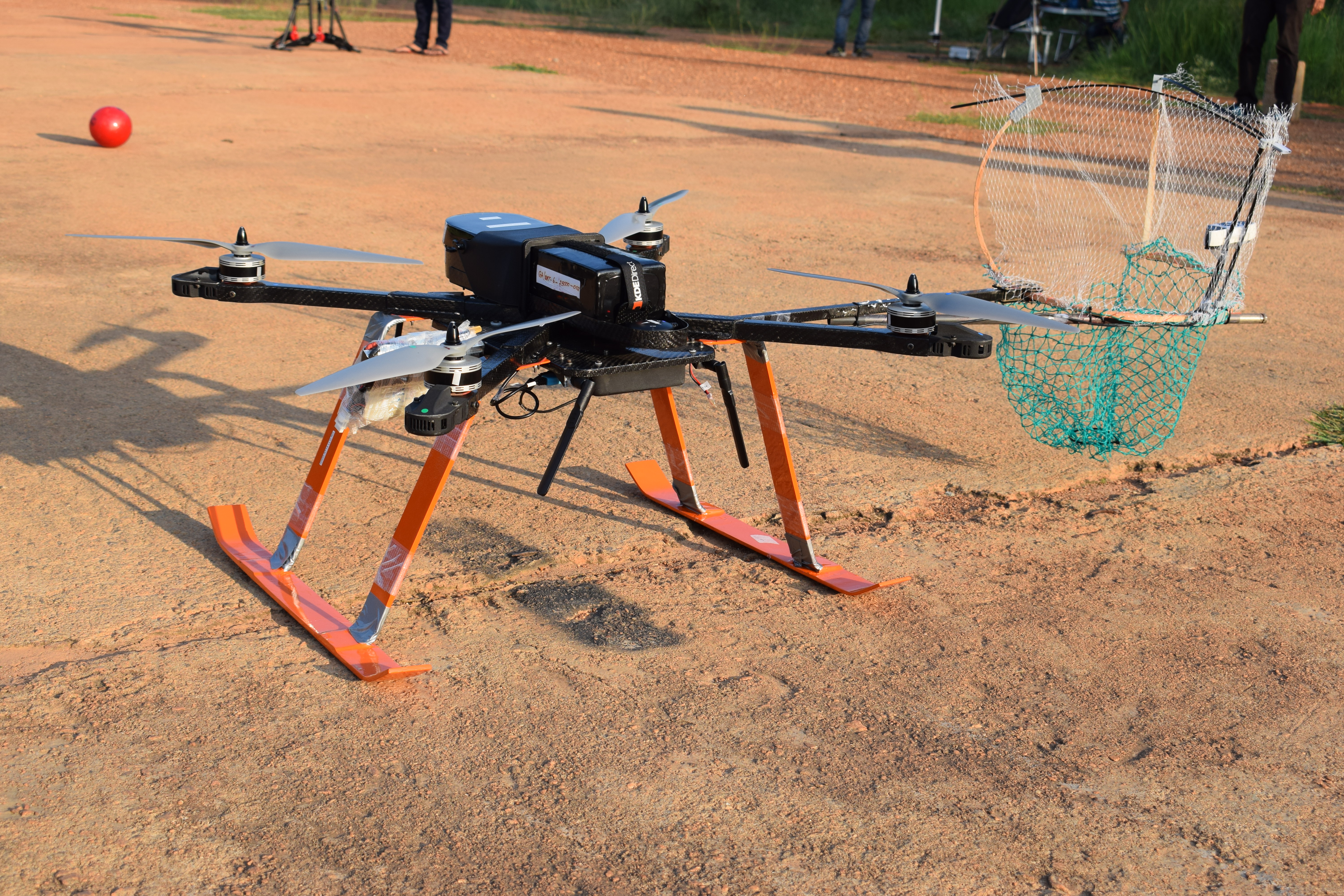}
    \caption{GA drone with the manipulator set-up}
    \label{drone_design1}
\end{figure}

The passive basket manipulator was tested in GA drone, which is a custom made quad rotor. The dimensions of this drone was designed to fit the volume constraints. The manipulator is fixed to the drone without any actuation as shown in Fig. \ref{drone_design1}. The landing gears resemble skiing skis, which are designed to absorb shock upon landing. 
%%Ive added this from previous draft ok
The custom drone has a wheelbase of 820 mm. The motors are placed at the corners of a square of side 580 mm. The complete air frame was made of carbon fibre composites. The maximum take off weight of the UAV was 4 kg (inclusive of 1.5 kg payload). The autopilot used was Pixhawk 2 Cube with Here GPS module. The on board computer was Nvidia Jetson TX2 with Auvidea J130 carrier board.

\begin{figure}
    \centering
    \includegraphics[scale=0.5]{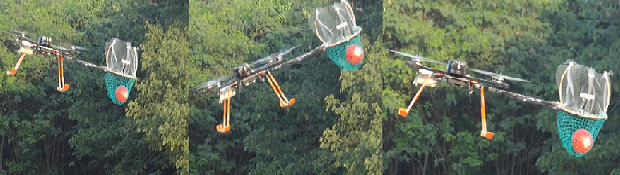}
    \caption{Oscillations observed in the roll plane of the drone due to moment created by weight imbalance}
    \label{fig:Ga_oscl}
\end{figure}

But the GA drone presented another problem. The manipulator acted as a cantilever beam with the end-effector and the captured ball as the end load. The moment generated resulted in oscillations in the roll plane, eventually leading to failure in grabbing. An instance of such an oscillation generated due to the moment imbalance, is shown in Fig. \ref{fig:Ga_oscl}. 
The moment generated is calculated as  $M = F \times d$, where, $F$ is the force and $d$ is the perpendicular distance from the point of interest (CG) to the line of action of the force. Hence, the moment generated by the end-effector is $M_{\text{arm}}= 1.2\times9.81\times0.91 =10.713 ~\text{Nm}$. The maximum take-off weight of the quadrotor was 4 kg and the generated thrust was 6 kg from the four motors. To counter the manipulator moment, the two motors towards the manipulator had to generate enough moment to keep the drone stable. Each of the motors generated a maximum thrust of 1.5 kg, and were situated at a distance of 0.290 m from the drone's centre of gravity (C.G.). The moment generated by the two motors together is $M_{\text{drone}}= 2\times1.5\times9.81\times0.29 =8.535$ Nm. The moment generated by the drone was insufficient to counter the moment created by the side ward load and the result was a continuous oscillation in the roll plane, observed throughout the mission.
Based on these tests, a larger drone was chosen which required the manipulator to have linear actuation for the manipulator, for the integrated system to remain within dimensional constraint.

\begin{figure}[htb!]
	\centering
	\includegraphics[width=1\linewidth]{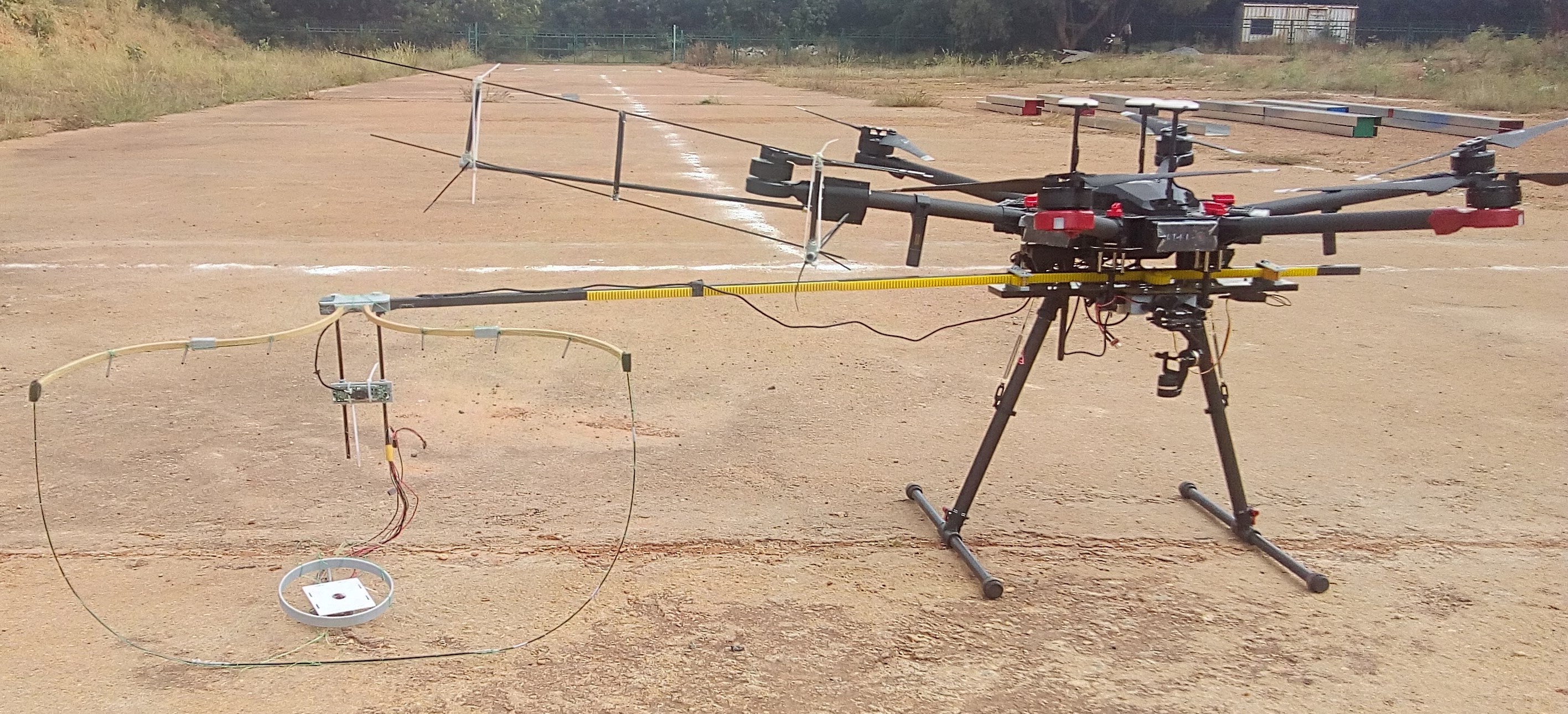}
    \caption{M600 hexa rotor with the passive basket set-up}
    \label{drone_design2}
\end{figure}

The M600 is a hexacopter with a maximum take-off weight of 15.5 kg. The thrust generated per motor is 2.5 kg placed at a distance of 0.570 m from the drone C.G.. There are three motors in this drone to counter the end-effector moment. The net counter moment generated is $M_{\text{M600}}=27.959$ Nm. The extension of the manipulator arm for the final design is 1.1 m. The motors are placed at the corners of a regular hexagon of side 580 mm. The air frame is made of carbon fibre composites and aluminium. The maximum take off weight of the UAV is 15.5 kg. The on-board autopilot is DJI A3 Pro, with triple redundant GPS. The on board computer used is the Nvidia Jetson TX2, with the supporting control board Arduino Mega for actuation. The drone has good maneuvering capability and better stability in comparison to the quad rotor drones. Since the wheel base of the drone itself occupies the desired volume, the manipulator is attached to the drone with an actuation mechanism, as shown in Fig. \ref{drone_design2}.

\subsection{Propeller guard}
\begin{figure}[htb!]
	\centering
	\begin{subfigure}{1\columnwidth}
	    \centering
	    \includegraphics[width=1\linewidth]{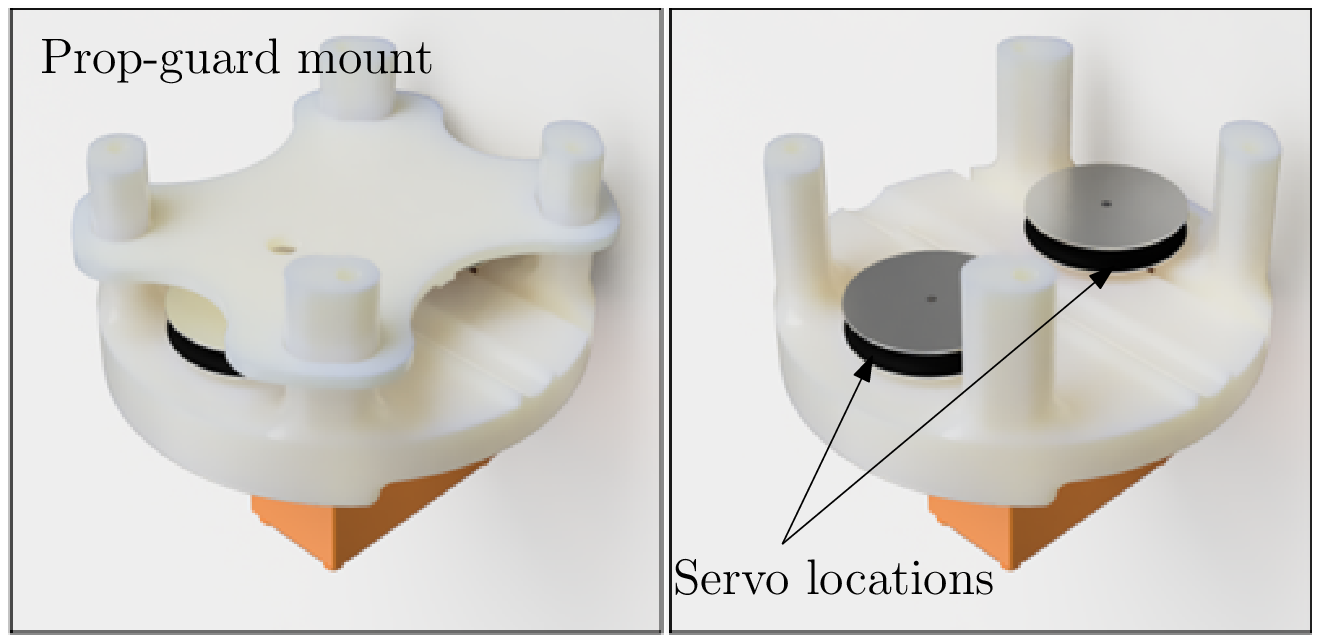}
	    \subcaption{}
	\end{subfigure}
	\begin{subfigure}{1\columnwidth}
	    \centering
	    \includegraphics[scale=0.45]{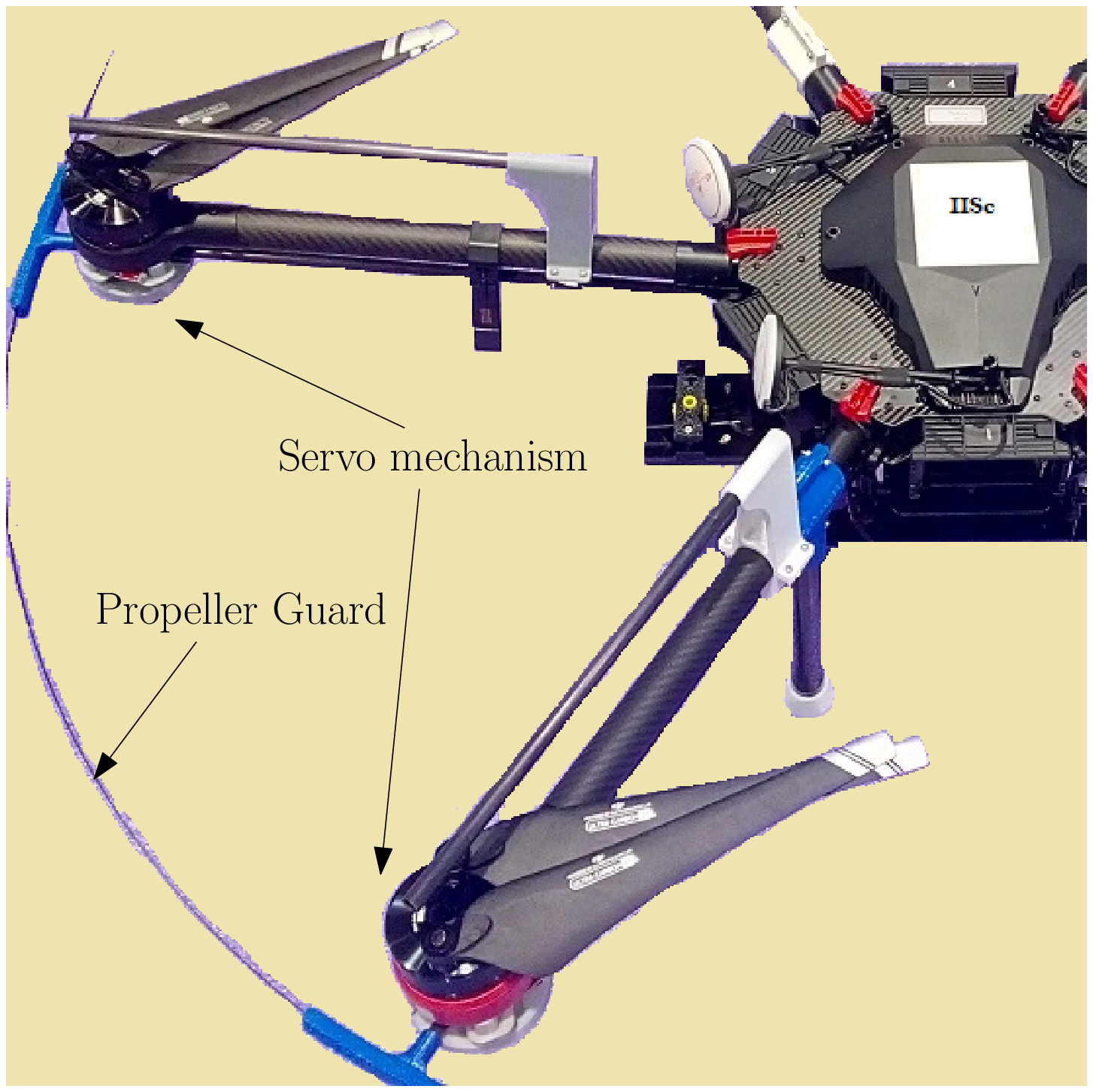}
	    \subcaption{}
	\end{subfigure}
	\begin{subfigure}{1\columnwidth}
	    \centering
	    \includegraphics[width=1\linewidth]{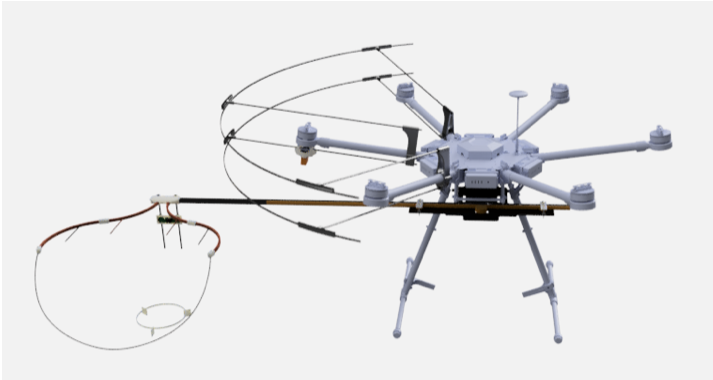}
	    \subcaption{}
	\end{subfigure}
	\caption{(a) Model of the propeller guard actuator mounts (b) Propeller guard prototype (c) Overall integrated system }
    \label{propguard-act}
\end{figure}
The objective of the entire task is to intercept the target ball and capture it while ensuring the safety of the drone. Since there is a possibility that the ball, the rod, or even the drone, could strike the drone's body or the propeller causing crashes, minimum safety has to be ensured. Since the manipulator was mounted sideways, the drone propellers on that side are susceptible to damage. To ensure the safety of the propellers on the manipulator side, an extendable propeller guard was designed, fabricated, and installed. The guard had two 3D printed standoffs on each of the three arms, that supported carbon fibre guide rods. Through these rods, inner telescopic rods extended to deploy the safety circumference around the three propellers. The extension gave sufficient clearance to the propellers. Actuation was based on friction sliding using a continuous rotation servo. The mechanism is optimally designed to be fixed under the drone motors without affecting the operation. The actuator set-up, the prototype, and the CAD model of the integrated system are shown in Fig. \ref{propguard-act}(a)-(c). 

\section{Discussion}\label{sec:7}
\begin{figure}[htb!]
	\centering
	\includegraphics[width=1\linewidth]{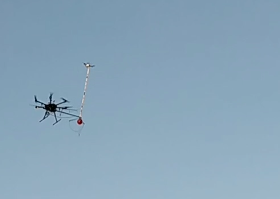}
    \caption{Ball grabbing using the proposed design}
    \label{grab}
\end{figure}

The prototype of manipulation mechanism designed is passive or energy efficient, could carry up to 150 g and can grab any object with size equivalent to a sphere of radius 20 cm. The prototype is made of wood and carbon fiber which provides good strength for multiple detachments. The entire end-effector assembly contributes to less drag and robust to mild wind disturbances. A snapshot from field test at IISc airfield, is shown in Fig. \ref{grab}.
After testing the prototype, few additional observations are noted below, which could be incorporated for better performance of future versions.
\begin{enumerate}
    \item A more detailed analysis of the prong geometry could reduce end-effector footprint while improving success of grabbing. 
    \item The additional detachment points were added based on tests, but a study on effect of position of these structures on grabbing success can help reduce improper placement that could adversely affect the performance for an alternative application
    \item The grab detector was made with 3 limit switches but a more energy efficient alternative could be achieved using piezo-electric sensors, which do not require external energy.
    \item The carbon fibre tube used for the manipulator arm was a standard square tube. Use of a rectangular or 'I' section tubes could add more rigidity while still being light-weight.
    \item The actuated propeller guard, though a novel design, was a preliminary effort towards ensuring safety. A more focused effort can made to make this system more effective with fewer components.
\end{enumerate}
\section{Conclusions}\label{sec:8}
This paper reported in detail, the design iterations followed to develop a novel passive basket-type manipulator. The challenges associated with the problem of aerial grabbing of maneuvering target, are addressed using the proposed design. The proposed design is compact and efficient for moving and low maneuvering targets. The design of novel guard is also presented which ensures safe operation of the drone. The maximum size of the object that can be grabbed is dependant on the size of the end-effector. The energy efficiency of the overall design and modular nature allowing portability, makes the design more attractive for different applications. 

\section*{Acknowledgements}
 We acknowledge all members of GCDSL and other team mates of Team IISc-TCS, for their contributions and suggestions towards this work. We acknowledge our collaborator, Tata Consultancy Services for their contributions in various directions towards this work.
\bibliographystyle{IEEEtran.bst}
\bibliography{References}

\end{document}